\documentclass[10pt,twocolumn,letterpaper]{article}

\usepackage{cvpr}              
\usepackage{booktabs}
\usepackage{multirow}
\usepackage{array}
\usepackage{xcolor}
\usepackage[accsupp]{axessibility}
\definecolor{cvprblue}{rgb}{0.21,0.49,0.74}
\usepackage[pagebackref,breaklinks,colorlinks,allcolors=cvprblue]{hyperref}

\def\paperID{21} 
\def\confName{CVPR}
\def\confYear{2026}

\title{DCMorph: Face Morphing via Dual-Stream Cross-Attention Diffusion}

\author {
    Tahar Chettaoui\textsuperscript{\textrm 1},
    Eduarda Caldeira\textsuperscript{\textrm 1, \textrm 3},
     Guray Ozgur\textsuperscript{\textrm 1}, 
     Raghavendra Ramachandra\textsuperscript{\textrm 2},\\
     Fadi Boutros\textsuperscript{\textrm 1},
    Naser Damer\textsuperscript{\textrm 1, \textrm 3} \\
    \textsuperscript{\rm 1}Fraunhofer Institute for Computer Graphics Research IGD, Germany\\
    \textsuperscript{\rm 2}Norwegian University of Science and Technology (NTNU), Norway \\
    \textsuperscript{\rm 3} TU Darmstadt, Germany
    \\
    {\tt\small tahar.chettaoui@igd.fraunhofer.de}
}

\begin{document}
\maketitle
\begin{abstract}
Advancing face morphing attack techniques is crucial to anticipate evolving threats and develop robust defensive mechanisms for identity verification systems. This work introduces DCMorph, a dual-stream diffusion-based morphing framework that 
simultaneously operates at both identity conditioning and latent space levels. Unlike image-level methods suffering from blending artifacts or GAN-based approaches with limited reconstruction fidelity, DCMorph leverages identity-conditioned latent diffusion models through two mechanisms: (1) decoupled cross-attention interpolation that injects identity-specific features from both source faces into the denoising process, enabling explicit dual-identity conditioning absent in existing diffusion-based methods, and (2) DDIM inversion with spherical interpolation between inverted latent representations from both source faces, providing geometrically consistent initial latent representation that preserves structural attributes. Vulnerability analyses across four state-of-the-art face recognition systems demonstrate that DCMorph achieves the highest attack success rates compared to existing methods at both operational thresholds, while remaining challenging to detect by current morphing attack detection solutions \url{https://github.com/TaharChettaoui/DCMorph}.
\end{abstract}
\vspace{-5mm}    
\section{Introduction}
\label{sec:intro}

Face recognition (FR) systems, despite their high accuracy, remain vulnerable to face morphing attacks, which create images that can be verified as belonging to multiple identities. Such attacks pose significant security threats in identity verification scenarios, potentially enabling unauthorized access for multiple people with the same document \cite{DBLP:conf/icb/FerraraFM14}. Understanding and developing advanced morphing techniques is essential to anticipate evolving attacks and strengthen defensive mechanisms against increasingly sophisticated threats.

Existing morphing approaches operate at two levels: image-level or representation-level. Image-level methods interpolate facial landmarks and blend textures, achieving strong identity preservation but suffering from visible blending artifacts \cite{DBLP:conf/icb/RaghavendraRVB17,DBLP:journals/tifs/FerraraFM18}. Representation-level methods, primarily GAN-based \cite{DBLP:conf/btas/DamerS0K18,DBLP:journals/tbbis/ZhangVRRDB21}, avoid such artifacts but exhibit limited reconstruction fidelity due to constrained latent space capacity. Recent diffusion-based morphing methods \cite{DBLP:conf/iwbf/DamerFSKHB23,DBLP:journals/tbbis/BlasingameL24}  achieves higher visual fidelity through latent space interpolation but operates without explicit identity-aware conditioning during the generation process, limiting its ability to precisely control identity characteristics in the synthesized morphs.

The emergence of identity-conditioned latent diffusion models (DM) \cite{DBLP:conf/iccv/BoutrosGKD23,DBLP:conf/iclr/LinHXMZD25}, which leverage cross-attention mechanisms to inject identity information into the denoising process, combined with DDIM inversion \cite{DBLP:conf/iclr/SongME21} that enables high-fidelity latent recovery, provides an opportunity to fundamentally advance face morphing. These developments enable simultaneous manipulation at both the conditioning pathway and the latent space, facilitating more effective identity blending while maintaining structural consistency.

This work introduces DCMorph, a dual-stream morphing framework that extends identity-conditioned DMs for face morphing. DCMorph combines decoupled cross-attention interpolation, which merges identity-specific features from both source faces during the denoising process, with spherical interpolation of DDIM-inverted latents, which provides geometrically consistent initialization for the denoising process. Our main contributions are:
\begin{itemize}
    \item A novel dual-stream morphing framework that simultaneously leverages identity conditioning via decoupled cross-attention and latent space interpolation via DDIM inversion, enabling more effective identity blending than existing approaches.
    \item Comprehensive vulnerability and detectability analyses demonstrating that DCMorph produces morphing attacks that are very effective against state-of-the-art FR systems in comparison to a set of both traditional and recent diffusion-based morphing methods, while posing challenging detectability characteristics.
\end{itemize}
\begin{figure*}[t]
    \centering
    \includegraphics[width=0.95\linewidth]{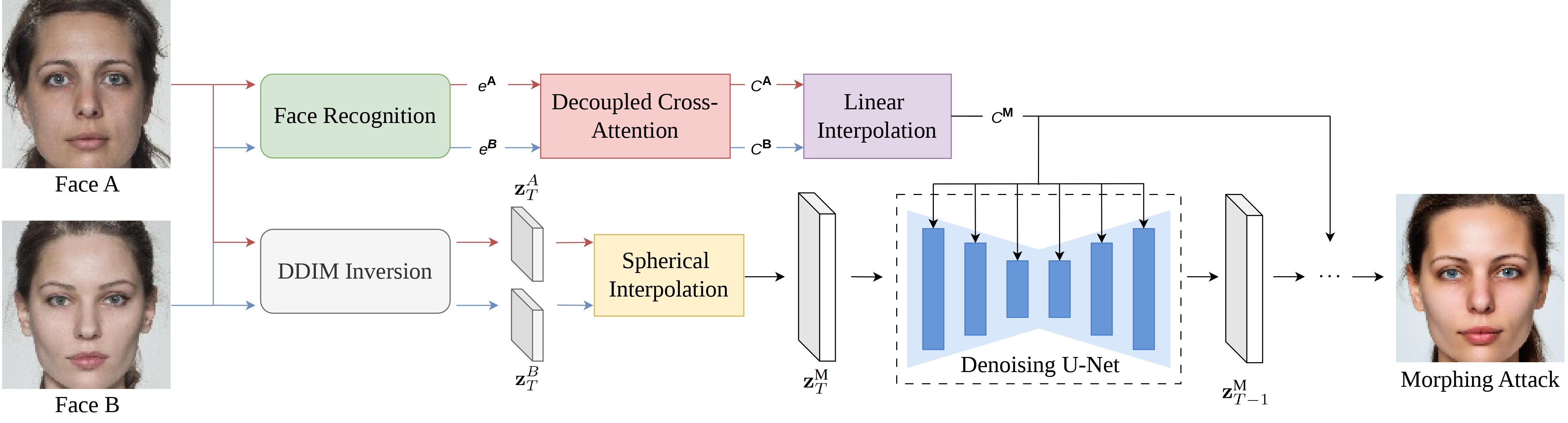}
    \vspace{-2mm}
    \caption{Overview of the DCMorph dual-stream framework. Identity embeddings $\mathbf{e}^A$ and $\mathbf{e}^B$ are extracted from input faces and injected via decoupled cross-attention layers to produce $\mathbf{C}^A$ and $\mathbf{C}^B$, which are linearly interpolated to obtain $\mathbf{C}^{\text{M}}$ (Eq. \ref{eq:ca_interpolation}). Simultaneously, DDIM inversion recovers latent representations $\mathbf{z}_T^A$ and $\mathbf{z}_T^B$, which are spherically interpolated to produce $\mathbf{z}_T^{\text{M}}$ (Eq. \ref{eq:spherical_interpolation}). The U-Net then performs iterative denoising ($\mathbf{z}_T^{\text{M}} \rightarrow \mathbf{z}_{T-1}^{\text{M}} \rightarrow \cdots \rightarrow \mathbf{z}_0^{\text{M}}$) conditioned on $\mathbf{C}^{\text{M}}$, generating a high-fidelity morphing attack blending identity characteristics from both sources through dual-stream manipulation at the conditioning and latent space levels.}
    \label{fig:approach}
    \vspace{-5mm}
\end{figure*}

\section{Related Work}
\label{sec:rw}

\textbf{Morphing Attack Generation Methods:}
Face morphing attacks create images verifiable as multiple identities, posing security threats to face recognition systems \cite{DBLP:conf/icb/FerraraFM14}. Existing approaches are categorized as image-level or representation-level, each with distinct characteristics. \textbf{Image-level Morphing:}
The earliest morphing attacks were created by detecting facial landmarks in source images, interpolating these landmarks, and blending texture information. Landmark-based morphs (LMA) \cite{DBLP:conf/icb/RaghavendraRVB17,DBLP:journals/tifs/FerraraFM18} have been widely studied, with early comparisons \cite{DBLP:journals/iet-bmt/ScherhagKRB20} showing certain approaches \cite{DBLP:conf/icb/RaghavendraRVB17,DBLP:conf/dagm/DamerBWBTBK18} achieved strong identity preservation. More advanced techniques perform partial-region interpolation \cite{DBLP:journals/tbbis/QinPVRLB21}, producing morphs that are more challenging to detect. Despite their effectiveness, image-level morphs inherently suffer from blending artifacts due to pixel-level manipulation and geometric inconsistencies \cite{DBLP:journals/tbbis/ZhangVRRDB21}. \textbf{Representation-level Morphing:}
To overcome these limitations, representation-level approaches perform interpolation in learned latent spaces. MorGAN \cite{DBLP:conf/btas/DamerS0K18} pioneered this direction, later enhanced through cascaded quality improvement \cite{DBLP:conf/btas/DamerBSKK19}. Building on this foundation, StyleGAN-based approaches \cite{DBLP:conf/iwbf/VenkateshZRRDB20} and MIPGAN I/II \cite{DBLP:journals/tbbis/ZhangVRRDB21} advanced the field through identity-preserving losses. However, GAN-based methods still suffer from limited reconstruction fidelity due to constrained latent capacity \cite{DBLP:conf/cvpr/PreechakulCWS22}, causing synthetic artifacts \cite{DBLP:journals/tbbis/ZhangVRRDB21}. More recently, diffusion-based morphing has demonstrated improved fidelity, with MorDIFF \cite{DBLP:conf/iwbf/DamerFSKHB23} and Blasingame and Liu \cite{DBLP:journals/tbbis/BlasingameL24} performing interpolation in diffusion latent spaces to achieve superior reconstruction quality. Beyond core generation techniques, post-processing methods have been developed to improve morph realism. ReGenMorphs \cite{DBLP:conf/isvc/DamerRSVBFKRK21} combines image-level morphing with GAN refinement, while Borghi et al. \cite{DBLP:journals/access/BorghiFGM21} and Di Domenico et al. \cite{DBLP:conf/iwbf/DomenicoBFM24} proposed attention-based and face restoration techniques for artifact retouching. Exploring alternative domains, Singh and Ramachandra \cite{DBLP:journals/tbbis/SinghR24,DBLP:conf/fgr/SinghR24} investigated 3D morphing via geometric interpolation and non-rigid registration, while MorCode \cite{DBLP:conf/bmvc/NRRM24a} introduced codebook-based morphing using discrete latent representations. Closely related to morph generation are demorphing methods, which attempt to recover the contributing identities from a morphed image. Shukla and Ross \cite{DBLP:conf/icb/ShuklaR24} proposed identity-preserving decomposition using generative frameworks, later introducing dc-GAN \cite{DBLP:conf/fgr/ShuklaR25} for dual-conditioned reconstruction of individual source identities.

\textbf{Diffusion Models for Face Generation:}
Diffusion probabilistic models (DPMs) \cite{DBLP:conf/nips/HoJA20,DBLP:conf/icml/Sohl-DicksteinW15} achieve superior visual fidelity over GANs by reversing gradual noising. Building on this, Latent Diffusion Models (LDMs) \cite{Rombach2021} operate in compressed latent space for efficient high-resolution generation, with U-Net denoising networks learning to remove noise conditioned on text, class labels, or identity embeddings. \textbf{Identity-Conditioned Diffusion:}
Recent methods \cite{DBLP:conf/iccv/BoutrosGKD23, DBLP:conf/nips/Xu0WXDJHM0DH24, DBLP:conf/cvpr/Kim00023, DBLP:conf/iclr/LinHXMZD25, caldeira2025negfacediff} inject identity embeddings from pre-trained FR models into denoising networks via cross-attention, enabling dynamic identity adaptation and high-fidelity identity-consistent synthesis. \textbf{Diffusion-based Morphing:}
Leveraging these advances, diffusion autoencoders \cite{DBLP:conf/cvpr/PreechakulCWS22} encode semantic and stochastic information for meaningful latent manipulations. MorDIFF \cite{DBLP:conf/iwbf/DamerFSKHB23} exploits this by interpolating these codes for high-fidelity morphing attacks. However, existing diffusion morphing performs only latent interpolation without explicit identity-aware conditioning during denoising.

DCMorph extends this work through dual-stream processing: decoupled cross-attention for dual-identity conditioning and DDIM inversion \cite{DBLP:conf/iclr/SongME21} with spherical interpolation for geometrically consistent latent initialization, enabling effective identity blending with high visual fidelity.

\vspace{-2mm}


\vspace{-1mm}
\section{Methodology} \label{sec:method}
\vspace{-1mm}
This section presents Dual-Stream Cross-Attention Morphing, a dual-identity conditioning framework for controlled identity morphing built upon pre-trained identity-conditional latent diffusion models (LDMs).

We begin this section by formalizing the LDM denoising process and the standard approach to identity-conditioned generation via cross-attention injection, which serves as the generative backbone of our morphing framework. We then introduce our two complementary morphing streams: Decoupled Identity Cross-Attention Interpolation, which injects identity-specific attention features from both source faces within the denoising network, U-Net, of LDM to guide the conditional generation toward a morphed identity representation, and Spherical Inverted Latents Interpolation, which applies DDIM inversion to each source image and performs spherical linear interpolation between the recovered latents to provide a geometrically consistent initialization for reverse diffusion. An overview of the full pipeline is illustrated in Figure \ref{fig:approach}.

\vspace{-1mm}
\subsection{Latent DM Preliminaries}
\label{sec:ldm}
\vspace{-1mm}

LDM \cite{Rombach2021} improves the efficiency of diffusion-based image generation by operating in a compressed latent space rather than directly in pixel space. Given a training image \( \mathbf{x}_0 \in \mathbb{R}^{H \times W \times C} \), a pre-trained encoder \( \mathcal{E}: \mathbb{R}^{H \times W \times C} \rightarrow \mathbb{R}^{d} \) maps it to a latent representation \( \mathbf{z}_0 = \mathcal{E}(\mathbf{x}_0) \). 

To learn the generative process, LDMs adopt the denoising diffusion probabilistic model (DDPM) framework \cite{DBLP:conf/nips/HoJA20}, where the latent code \( \mathbf{z}_0 \) is gradually perturbed by Gaussian noise over \( T \) discrete timesteps:
\begin{equation}
q(\mathbf{z}_t | \mathbf{z}_{t-1}) = \mathcal{N}(\mathbf{z}_t; \sqrt{1 - \beta_t} \mathbf{z}_{t-1}, \beta_t \mathbf{I}),
\end{equation}
where \( \{ \beta_t \}_{t=1}^T \) is a predefined variance schedule. After sufficient noising steps, the latent \( \mathbf{z}_T \) approximates an isotropic Gaussian distribution, i.e., \( \mathbf{z}_T \sim \mathcal{N}(0, \mathbf{I}) \).

The reverse process is modeled by a neural network \( \theta \), typically implemented as a U-Net \cite{Rombach2021}, which is trained to denoise the latent variable by predicting the noise component added at each step. The network takes as input the noisy latent \( \mathbf{z}_t \), the timestep \( t \), and an optional conditioning signal \( \mathbf{c} \in \mathbb{R}^d \), such as an identity embedding from a pretrained face recognition model. The reverse process is thus formulated as:
\begin{equation}
\label{eq:denoise}
p_\theta(\mathbf{z}_{t-1} | \mathbf{z}_t, \mathbf{c}) = \mathcal{N}(\mathbf{z}_{t-1}; \mu_\theta(\mathbf{z}_t, t, \mathbf{c}), \Sigma_\theta(\mathbf{z}_t, t, \mathbf{c})).
\end{equation}

After the denoising trajectory is complete, the final latent \( \hat{\mathbf{z}}_0 \) is passed through a pre-trained decoder \( \mathcal{D}: \mathbb{R}^{d} \rightarrow \mathbb{R}^{H \times W \times C} \), to reconstruct image \( \hat{\mathbf{x}}_0 = \mathcal{D}(\hat{\mathbf{z}}_0) \).

\textbf{Classifier-Free Guidance (CFG):}
Classifier-Free Guidance (CFG)~\cite{DBLP:journals/corr/abs-2207-12598} is applied to reinforce the impact of the condition (e.g., identity) during sampling~\cite{DBLP:conf/iclr/LinHXMZD25}. It is effective in a wide range of conditional generation tasks, including both image- and text-conditioned synthesis~\cite{ban2024understanding, DBLP:conf/icml/Wang0HG24, DBLP:conf/iclr/LinHXMZD25}.

At inference time, the final noise prediction is obtained by linearly combining the conditional and unconditional outputs:
\begin{equation}
\label{eq:cfg}
\hat{\epsilon} = (1 + \omega)\hat{\epsilon}_\theta(\mathbf{z}_t, t, \mathbf{c}) - \omega \hat{\epsilon}_\theta(\mathbf{z}_t, t),
\end{equation}
where $\omega$ denotes the guidance strength. Increasing $\omega$ amplifies the contribution of the identity-conditioned prediction, leading to stronger adherence to the specified identity and improved identity consistency in the generated samples.

\textbf{Identity-conditional LDMs:} 
To generate identity-conditioned face images using an LDM, the common approach \cite{DBLP:conf/iccv/BoutrosGKD23, DBLP:conf/nips/Xu0WXDJHM0DH24, DBLP:conf/cvpr/Kim00023, DBLP:conf/iclr/LinHXMZD25} is to utilize identity representation \( \mathbf{c} \) extracted from a pretrained FR model as a condition for LDM.
The identity representation \( \mathbf{c} \) is injected into the U-Net via cross-attention layers \cite{Rombach2021}. This mechanism projects the identity context from a single source into the denoising network's intermediate feature representations, allowing the generative process to adapt dynamically to the given identity \cite{Rombach2021, DBLP:conf/iccv/BoutrosGKD23}. In contrast to single-identity conditioning, identity morphing requires simultaneous conditioning on two source images, enabling the generative process to synthesize a representation that integrates identity characteristics from both inputs. The next section describes our morphing approach that injects identity-representation from two source images into the denoising process.

\begin{figure*}[h!]
    \centering
    \includegraphics[width=0.8\textwidth]{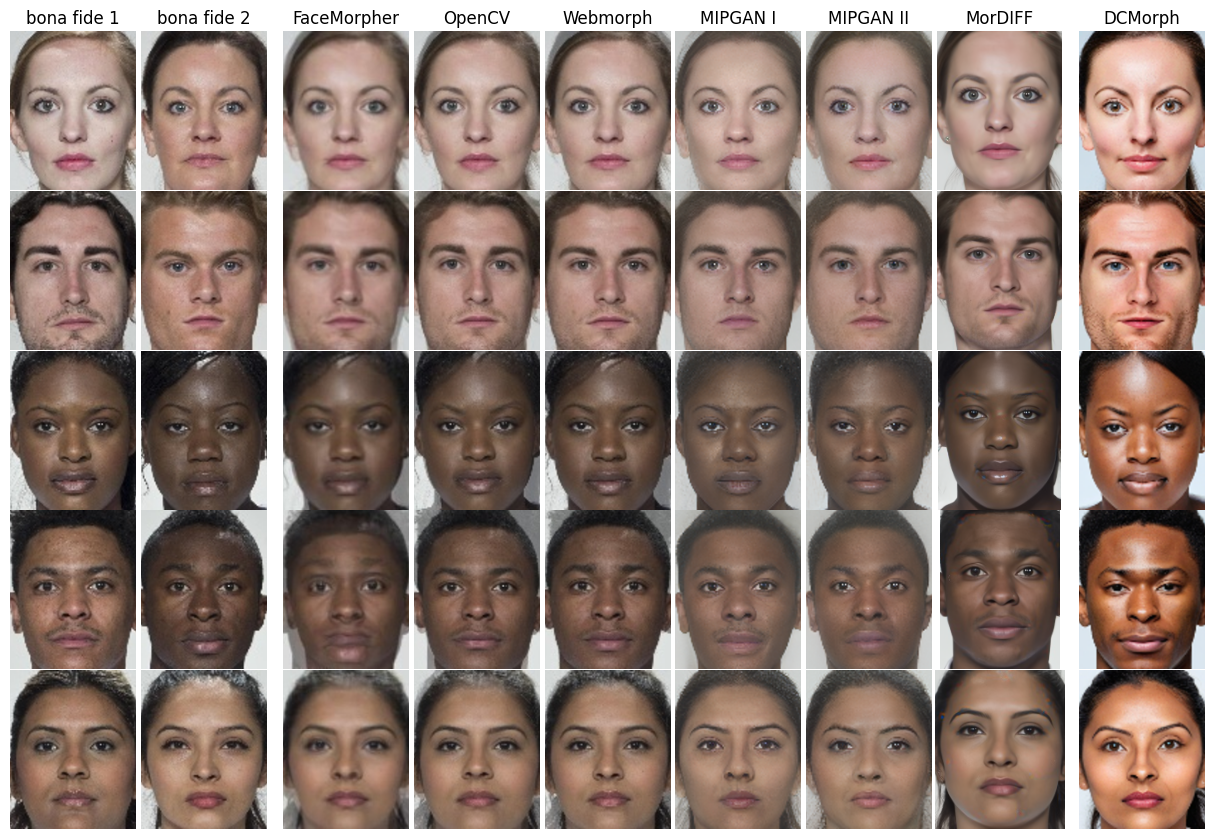}
    \vspace{-2mm}
    \caption{Samples of the DCMorph (right most column) attacks, the baseline attacks (created by FaceMorpher, OpenCV, WebMorph, MorDIFF, MIPGANI and II), and the bona fide images that were morphed to creat the attack (two left-most columns). The image-level morphs (FaceMorpher, OpenCV, WebMorph) show the traditional blending artifacts, while the representation-level morphs (MIPGAN-I and II) show typical streaking GAN artifacts. Diffusion-based approaches, such as ours and MorDIFF, exhibit substantially fewer generative artifacts compared to GAN-based methods.}
    \label{fig:sample}
    \vspace{-5mm}
\end{figure*}

\subsection{DCMorphing: Dual-Stream Cross-Attention Morphing} \label{sec:dcmorph}
This section introduces \textbf{Dual-Stream Cross-Attention Morphing (DCMorphing)}, a dual-stream morphing framework that extends identity-conditioned LDMs by jointly combining identity information in both the conditioning pathway and the latent space. The two streams operate at complementary levels: cross-attention interpolation merges identity features during denoising, while latent interpolation defines a geometrically meaningful generative trajectory.


\textbf{Decoupled Identity Cross-Attention Interpolation:}
Let \( \mathbf{x}^A, \mathbf{x}^B \) denote two input face images to be morphed. Identity embeddings are extracted as:
\begin{equation}
\mathbf{c}^A = f(\mathbf{x}^A), \quad
\mathbf{c}^B = f(\mathbf{x}^B).
\end{equation}
Each embedding is injected into the denoising network, U-Net, via decoupled cross-attention layers.
For identity \( A \), the output of the cross-attention is defined as:
\begin{equation}
\mathbf{C}^A = \text{Attn}(Q, K^A, V^A) = \text{softmax}\Big(\frac{Q (K^A)^\top}{\sqrt{d}}\Big)V^A,
\end{equation}

where the query ($Q$), key ($K$), and value ($V$) matrices are computed as $Q = F W_q$, $K^A = \mathbf{c}^A W_k$, and $V^A = \mathbf{c}^A W_v$, with $F$ denoting the query features from the U-Net, and $W_q, W_k, W_v$ being the weight matrices of the trainable linear projection layers. 

Similarly, we compute $\mathbf{C}^B = \text{Attn}(Q, K^B, V^B)$ using its respective identity embedding $\mathbf{c}^B$:
\begin{equation}
\mathbf{C}^B = \text{Attn}(Q, K^B, V^B) = \text{softmax}\Big(\frac{Q (K^B)^\top}{\sqrt{d}}\Big)V^B.
\end{equation}
Noting that the $W_q, W_k, W_v$  weights are shared between the two cross-attention layers.
These cross-attention outputs are then linearly interpolated to produce a mixed identity representation for downstream generation. We denote linear interpolation as $r_{l}(\lambda; A, B)$, where $\lambda \in [0,1]$ is the interpolation coefficient between the anchors $A$ and $B$. The final formulation of the decoupled interpolated cross-attention $\mathbf{C}^{\text{M}}$ is therefore defined as:
\begin{equation}
\label{eq:ca_interpolation}
\mathbf{C}^{\text{M}} = r_{l}(\lambda; \mathbf{C}^A, \mathbf{C}^B)  = \lambda \mathbf{C}^A + (1-\lambda) \mathbf{C}^B,
\end{equation}

where $\lambda$ controls the relative contribution of the two identity-specific cross-attention outputs. In our experiments, we set $\lambda = 0.5$, assigning equal weight to both components. As a result, the synthesized representation reflects an equal contribution of identity-specific features from both sources. This balanced interpolation in the attention space encourages the generated output to retain characteristic attributes of each identity, while ensuring structural and semantic consistency in the resulting morph.

\textbf{Spherical Inverted Latents Interpolation:}
In addition to performing decoupled identity cross-attention interpolation, we also operate in the latent space by applying DDIM inversion \cite{DBLP:conf/iclr/SongME21} to each input image to recover their corresponding latent representations, and subsequently perform spherical interpolation between these latents to generate an intermediate representation that effectively captures a balanced combination of the original face images. For input face images $\mathbf{x}^A$ and $\mathbf{x}^B$ corresponding to identities $A$ and $B$, we denote the DDIM inversion process by $d(\cdot)$, which maps an image to its latent representation by iteratively adding noise:

\begin{equation}
\mathbf{z}_T^A = \text{d}(\mathbf{x}^A), \quad
\mathbf{z}_T^B = \text{d}(\mathbf{x}^B), \quad 
\mathbf{z}_T^A, \mathbf{z}_T^B \sim \mathcal{N}(0, \mathbf{I}).
\end{equation}

Ideally, direct sampling from $\mathbf{z}_T^A$ and $\mathbf{z}_T^B$ without any further manipulation should reconstruct an image that closely approximates $\mathbf{x}^A$ and $\mathbf{x}^B$, respectively. To interpolate between the two identities in latent space, we perform spherical linear interpolation (slerp) \cite{DBLP:conf/eccv/JangHSCL24} between $\mathbf{z}_T^A$ and $\mathbf{z}_T^B$. Spherical interpolation is specifically employed to maintain the Gaussian statistics of the latent space \cite{DBLP:conf/nips/SamuelBDMC23}, ensuring that the interpolated latent $\mathbf{z}_T^\text{M}$ remains consistent with the distribution of valid latents. We denote spherical interpolation as $r_{s}(\lambda; A, B)$, respectively, where $\lambda \in [0,1]$ represents the interpolation coefficient, and $A$ and $B$ denote the interpolation anchors. With $\theta = \textit{arccos}\frac{ \mathbf{z}_T^A \cdot \mathbf{z}_T^B}{\|\mathbf{z}_T^A\| \|\mathbf{z}_T^B\|}$, the interpolated latent $\mathbf{z}_T^\text{M}$ is defined as:
\begin{equation}
\label{eq:spherical_interpolation}
\mathbf{z}_T^{\text{M}} = r_{s}(\lambda; \mathbf{z}_T^A, \mathbf{z}_T^B)
= \frac{\sin(1-\lambda)\theta}{\sin \theta} \mathbf{z}_T^A + \frac{\sin(\lambda \theta)}{\sin \theta} \mathbf{z}_T^B,
\end{equation}

This procedure produces an intermediate latent representation that is intended to mix the identity-specific features of both inputs, providing a suitable starting point for subsequent generation or morphing. The reverse diffusion process is then applied to $\mathbf{z}_T^\text{M}$ to generate the corresponding interpolated image, which is expected to reflect characteristics from both identities in a balanced manner.

The interpolation strategies are designed to contribute distinct but synergistic benefits to the morphing process.  Interpolating identities via decoupled cross-attention interpolation aims to influence the model’s conditional feature aggregation, guiding the denoising process toward a representation that contains features from both identities during generation, consistent with how cross‑attention enables flexible conditioning in LDMs \cite{Rombach2021}. In addition, spherical interpolation in latent space preserving high‑level structural and semantic attributes encoded in the latent space, this approach provides a stable initialization for the diffusion process, providing a geometrically consistent trajectory in the generative prior, preserving global structure and semantic coherence. Collectively, these mechanisms aim to produce interpolated images that uphold both realistic generation quality and balanced representation of the source identities.

\vspace{-2mm}
\section{Experimental Setup} \label{sec:experimentalsetup} 
\vspace{-2mm}
\textbf{Morph Generation Protocol:} The DCMorph dataset extends over the SYN-MAD 2022 competition \cite{DBLP:conf/icb/HuberBLRRDNGSCT22} and uses the same morphing pairs to enable a comparable dataset. Both are based on the Face Research Lab London (FRLL) dataset \cite{DeBruine2021}. The FRLL contains images of 102 different individuals and provides high-quality frontal images created in a controlled scenario with a wide range of different ethnicities. All individuals present in the dataset signed consent for their images to be used in lab-based and web-based studies in their original or altered forms and to illustrate research. For the morph generation, we limited the data to the frontal images of the dataset, following SYN-MAD 2022. The pairs are defined in SYN-MAD 2022 by splitting the frontal images of the FRLL depending on the provided gender and expression (neutral or smiling). ElasticFace-Arc \cite{DBLP:conf/cvpr/BoutrosDKK22} was then used to generate embeddings of the images and these embeddings are then compared with cosine similarity within the split to find the most similar faces. The 250 most similar face pairs are selected, resulting in 250 female neutral pairs, 250 female smiling pairs, 250 male neutral pairs, and 250 male smiling pairs, for a total of 1000 attack images and the 204 bona fide images of the SYN-MAD 2022.

\textbf{Benchmark Datasets:} We use the SYN-MAD 2022 \cite{DBLP:conf/icb/HuberBLRRDNGSCT22} benchmark, containing morphed images from 5 different approaches, three image-level (FaceMorpher (commercial-of-the-shelf), OpenCV \cite{openCVmorph} and Webmorph (online tool \footnote{https://webmorph.org/} ) and two representation-level GAN-based (MIPGAN I \cite{DBLP:journals/tbbis/ZhangVRRDB21} and MIPGAN II \cite{DBLP:journals/tbbis/ZhangVRRDB21}). Additionally, we add the more recent MorDIFF \cite{DBLP:conf/iwbf/DamerFSKHB23} a representation-level diffusion-based approach. We follow the same morph pair selection protocol and add our DCMorph attacks to the benchmark, which will be publicly released.

\textbf{Vulnerability of FR systems:} We evaluated the vulnerability of four FR systems to the DCMorph attacks in comparison to six different attacks. The FR systems are AdaFace \cite{DBLP:conf/cvpr/Kim0L22}, ArcFace \cite{DBLP:conf/cvpr/DengGXZ19}, ElasticFace (ElasticFace-Arc) \cite{DBLP:conf/cvpr/BoutrosDKK22} and CurricularFace \cite{DBLP:conf/cvpr/HuangWT0SLLH20}. All considered models are based on ResNet-100 architectures and have $55.52$M parameters with $24192.51$ MFLOPs. All the FR models are the official releases by the respective authors. We present the vulnerability results by reporting the Mated Morphed Presentation Match Rate (MMPMR) \cite{DBLP:conf/biosig/ScherhagNRGVSSM17}, evaluated at decision thresholds corresponding to false match rates (FMR) of 1\% and 0.1\%, denoted as MMPMR100 and MMPMR1000, respectively. The FMR decision thresholds were calculated on the LFW \cite{huang:inria-00321923} benchmark, following \cite{DBLP:conf/icb/HuberBLRRDNGSCT22}. 

\textbf{Detectability of Morphing Attacks:} The MAD performance (detectability) is presented by the Attack Presentation Classification Error Rate (APCER), i.e. the proportion of attack images incorrectly classified as bona fide samples, at a fixed (1\%, 10\%, and 20\%) Bona fide Presentation Classification Error Rate (BPCER), i.e. the proportion of bona fide images incorrectly classified as attack samples, as defined in the ISO/IEC 30107-3 \cite{ISO30107-3-2016}. Additionally, the Detection Equal Error Rate (EER), i.e. the value of APCER or BPCER at the decision threshold where they are equal, is reported. We use three MAD systems to evaluate the morphing attack detectability of the proposed dataset. MADPromptS \cite{DBLP:journals/corr/abs-2508-08939} utilizes multiple prompt aggregation to exploit the full potential of zero-shot learning MAD in foundation models. SPL-MAD \cite{fang2022unsupervised} is an unsupervised method based on self-paced learning trained for anomaly detection. The public released supervised MixFaceNet-MAD \cite{damer2022privacy} uses the efficient MixFaceNet \cite{DBLP:conf/icb/BoutrosDFKK21} architecture (originally developed for FR) trained with the binary cross-entropy loss function to detect morphing attacks.

\textbf{Base Diffusion Model:} The generative backbone of DCMorph is built upon Stable Diffusion XL (SDXL) \cite{DBLP:journals/corr/abs-2307-01952}, fine-tuned via IP-Adapter \cite{ye2023ip-adapter} to support conditional face generation \footnote{\url{https://huggingface.co/h94/IP-Adapter-FaceID}}. Specifically, IP-Adapter extends the pre-trained SDXL model by introducing a lightweight adapter module that enables image-prompt conditioning via dual cross-attention layers, for text and image conditions, allowing face-identity embeddings to be injected into the denoising U-Net without modifying the original model weights. We eliminated the cross-attention layer of the text condition and duplicated the face identity condition layer, enabling dual conditioning from both source identities. The conditional face embeddings used to guide the generation processes are extracted from a pre-trained ArcFace \cite{DBLP:conf/cvpr/DengGXZ19} model. 

\begin{figure}[t] 
    \centering
    \includegraphics[width=\columnwidth]{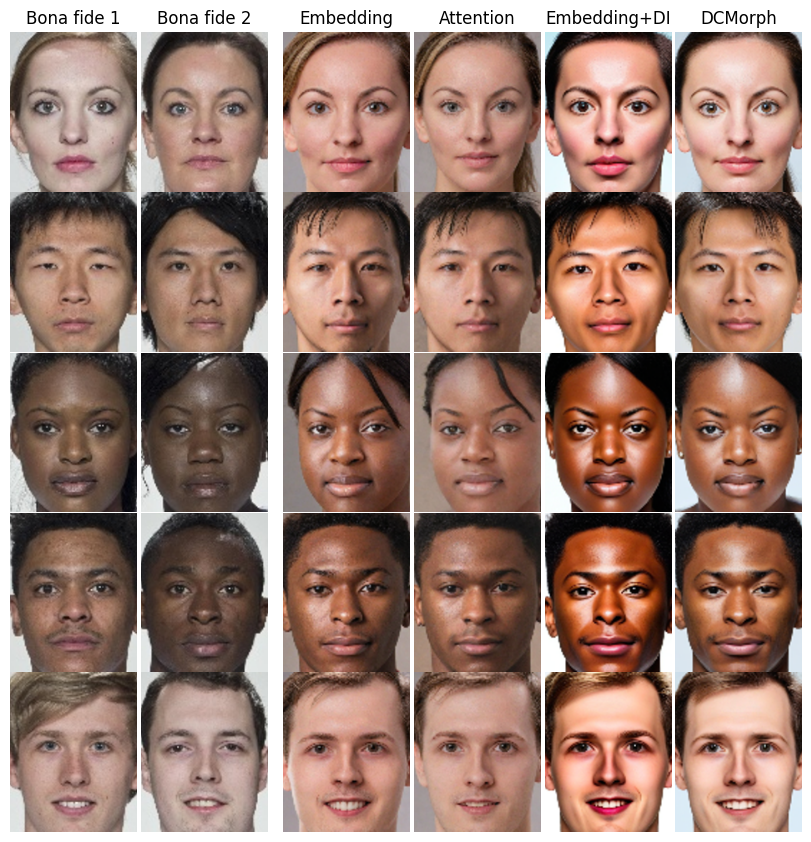}
    \caption{Qualitative comparison of morphing approaches. Bona fide images (left columns) and the corresponding DCMorph attacks (right column), generated using decoupled cross-attention identity interpolation combined with spherical latent-space interpolation via DDIM inversion, are shown alongside the alternative approaches considered in this work. DDIM inversion preserves high-level latent semantics, enabling a geometrically consistent reverse diffusion process that maintains global structure and attributes such as head pose.}
    \label{fig:ablation}
    \vspace{-5mm}
\end{figure}

\begin{table*}[t] 
\centering
\caption{Ablation study analyzing the vulnerability of four FR systems to different morphing strategies. The table compares embedding interpolation alone, cross-attention interpolation alone, embedding interpolation with DDIM inversion, and DCMorph (cross-attention with DDIM inversion). Embedding interpolation operates in the feature space of pre-trained FR models, cross-attention injects dual-identity conditioning into the denoising process, and DDIM inversion with spherical interpolation provides geometrically consistent latent initialization. DCMorph achieves the highest MMPMR values (bold) across all FR systems and both operational thresholds (FMR 1\% and 0.1\%), demonstrating the synergistic benefit of combining identity-conditioned generation with latent space interpolation.}
\vspace{-2mm}
\setlength{\tabcolsep}{6pt}
\renewcommand{\arraystretch}{1.2}
\resizebox{0.9\textwidth}{!}{%
\begin{tabular}{c c c ccc c c c c c c c}
\toprule
 \multirow{2}{*}{Morphing Technique}
& \multicolumn{3}{c}{Method}
& \multicolumn{2}{c}{ElasticFace \cite{DBLP:conf/cvpr/BoutrosDKK22}}
& \multicolumn{2}{c}{CurricularFace \cite{DBLP:conf/cvpr/HuangWT0SLLH20}}
& \multicolumn{2}{c}{AdaFace \cite{DBLP:conf/cvpr/Kim0L22}}  
& \multicolumn{2}{c}{ArcFace \cite{DBLP:conf/cvpr/DengGXZ19}} \\
\cmidrule(lr){2-4} \cmidrule(lr){5-6} \cmidrule(lr){7-8} \cmidrule(lr){9-10} \cmidrule(lr){11-12}
& Embedding Interpolation & Cross-Attention & DDIM-Inversion & MMPMR100 & MMPMR1000 & MMPMR100 & MMPMR1000 & MMPMR100 & MMPMR1000 & MMPMR100 & MMPMR1000 \\
\midrule
Embedding Interpolation & X &  &  & 0.992 & 0.936 & 0.995 & 0.960 & 1.000 & 0.986 & 0.998 & 0.966 \\
Cross-attention Interpolation &  & X &  & 0.990 & 0.916 & 0.992 & 0.956 & 0.999 & 0.978 & 0.993 & 0.963 \\
Embedding Interpolation + DDIM & X &  & X & 0.997 & 0.959 & 0.996 & 0.969 & 0.999 & 0.991 &  0.993 & 0.963\\
DCMorph &  & X & X & 1.000 & 0.965 & 0.999 & 0.981 & 1.000 & 0.995 & 0.998 & 0.982 \\
\bottomrule
\end{tabular}}
\label{tab:morph_results_ablation}
\vspace{-3mm}
\end{table*}

\vspace{-2mm}
\section{Results} \label{sec:results} 
\vspace{-2mm}

We first conduct an ablation study to systematically evaluate the individual and combined contributions of embedding interpolation, cross-attention interpolation, and DDIM inversion to morph generation effectiveness. The results show that both interpolation strategies alone already produce highly vulnerable morphs, while incorporating DDIM inversion further strengthens identity preservation under stricter verification settings. 
The full DCMorph framework consistently achieves the strongest vulnerability across all FR systems. From the detectability perspective, the findings reveal that although some intermediate variants are more easily detected by certain MAD approaches, the complete DCMorph pipeline significantly increases detection difficulty for the considered MAD approaches. We then compare DCMorph with established baseline attack methods, analyzing both vulnerability and detectability across multiple evaluation scenarios. Figure \ref{fig:sample} provides a qualitative comparison of DCMorph attacks with other morphing methods. The combined detectability and vulnerability results show that DCMorph achieves a favorable trade off by maximizing FR vulnerability while remaining difficult to detect. 
Even the zero shot foundation model MADPromptS demonstrates only moderate detection capability, with DCMorph remaining harder to detect than most attacks.

\vspace{-1mm}
\subsection{DCMorph Component Analysis} 
\vspace{-1mm}
In this section, we assess the vulnerability and detectability of different morphing strategies to understand the individual and combined contributions of DCMorph's components. We evaluate four variants: (1) embedding interpolation alone, (2) cross-attention interpolation alone, (3) embedding interpolation combined with DDIM inversion, and (4) DCMorph, which combines cross-attention interpolation with DDIM inversion. 

\textbf{Vulnerability Analysis:} 
Table \ref{tab:morph_results_ablation} presents the vulnerability of four state-of-the-art (SOTA) FR systems to different morphing techniques. Embedding interpolation alone achieves high MMPMR values across all FR systems, ranging from 0.992 to 1.000 at MMPMR100 and 0.936 to 0.986 at MMPMR1000, demonstrating that direct interpolation of face embeddings can already generate highly effective morphing attacks. Cross-attention interpolation performs comparably, with MMPMR100 values between 0.990 and 0.999, though showing slightly lower MMPMR1000 scores (0.916 to 0.978), suggesting minor effectiveness reductions at the more restrictive operational threshold. Combining embedding interpolation with DDIM inversion yields improvements across most FR models, demonstrating that spherical interpolation of inverted latents provides better identity preservation at stricter thresholds. However, DCMorph, which combines cross-attention interpolation with DDIM inversion, achieves the highest vulnerability across all evaluation settings. DCMorph scores perfect or near-perfect MMPMR100 values (0.998–1.000) and the highest MMPMR1000 values (0.965–0.995) for all four FR systems. Notably, on AdaFace and ElasticFace, DCMorph achieves perfect scores (1.000) at MMPMR100, while on CurricularFace and ArcFace, it reaches MMPMR100 values of 0.999 and 0.998 respectively, outperforming all other variants. The superior performance of DCMorph demonstrates the synergistic benefit of jointly operating at both the conditioning pathway and latent space levels. Figure \ref{fig:ablation} provides qualitative comparisons showing how DDIM inversion preserves high-level latent semantics, maintaining global structure and attributes more consistently than other approaches, such as slight head pose inconsistency that might be seen more clearly in the third row of Figure \ref{fig:ablation}.

\textbf{Detectability Analysis:} 
Table \ref{tab:mad_results_ablation} presents the detectability of different morphing techniques across three MAD systems, presented in Section \ref{sec:experimentalsetup}. The results reveal complex patterns depending on the MAD architecture and detection strategy. For MADPromptS \cite{DBLP:journals/corr/abs-2508-08939}, a zero-shot foundation model approach, morphing attacks generated through embedding interpolation and cross-attention interpolation alone are poorly detected, with EERs of 41.90\% and 44.00\% respectively, and APCER values exceeding 84\% at 10\% BPCER. In contrast, DCMorph shows improved detectability from the MAD perspective, with a lower EER of 27.60\% and APCER of 48.90\% at 10\% BPCER, though still representing a challenging detection scenario. For SPL-MAD \cite{fang2022unsupervised}, an unsupervised anomaly detection method, attacks generated with embedding interpolation remains moderately challenging to detect (EER 38.30\%), while cross-attention interpolation becomes substantially easier to detect. However, DCMorph becomes extremely difficult to detect by this MAD, achieving an EER of 77.60\% with APCER values of 100\% across all BPCER operating points. Similarly, for MixFaceNet-MAD \cite{damer2022privacy}, cross-attention interpolation alone is relatively well-detected (EER 6.40\%), while DCMorph becomes highly challenging to detect (EER 69.20\%, APCER 100\% at both 1\% and 10\% BPCER). These results demonstrate that while DCMorph's dual-stream approach maximizes FR vulnerability, it also creates detection challenges for current MAD solutions, particularly those based on supervised learning with limited exposure to such attacks.

\begin{table}[ht!] 
\centering 
\caption{Detectability analysis of different morphing strategies across three MAD systems. The table presents EER and APCER at three BPCER operating points (1\%, 10\%, 20\%). Lower EER and APCER indicate easier detection (better MAD performance). DCMorph shows varied detectability patterns: moderately challenging for MADPromptS, but extremely challenging for SPL-MAD and MixFaceNet-MAD. 
} 
\vspace{-2mm}
\resizebox{0.45\textwidth}{!}{
\begin{tabular}{llcccc}
\toprule
\multirow{2}{*}{Method} & \multirow{2}{*}{Test data}
& \multirow{2}{*}{EER (\%)}
& \multicolumn{3}{c}{APCER (\%) @ BPCER (\%)} \\
\cmidrule(lr){4-6}
& & & 1.00 & 10.00 & 20.00 \\
\midrule
\multirow{4}{*}{MADPromptS \cite{DBLP:journals/corr/abs-2508-08939}}
& Embedding Interpolation &  41.90 & 99.90 & 84.70 & 68.50 \\
& Cross-attention Interpolation & 44.00 & 99.50 &  84.20 & 71.70 \\
& Embedding Interpolation + DDIM & 32.40 & 96.30 & 60.40 & 44.60  \\
& DCMorph & 27.60 & 92.50 & 48.90 & 35.70 \\
\midrule
\multirow{4}{*}{SPL-MAD \cite{fang2022unsupervised}}
& Embedding Interpolation & 38.30 & 99.00 & 77.90 & 66.70\\
& Cross-attention Interpolation & 11.00 & 75.70 & 12.30 & 6.40  \\
& Embedding Interpolation + DDIM & 32.40 & 96.30 & 60.40 & 44.60 \\
& DCMorph & 77.60 & 100.0 & 100.0 & 100.0 \\
\midrule
\multirow{4}{*}{MixFaceNet-MAD \cite{damer2022privacy}}
& Embedding Interpolation & 17.50 & 99.70 & 38.20 & 11.60 \\
& Cross-attention Interpolation & 6.40 & 85.40 & 3.70 & 0.30 \\
& Embedding Interpolation + DDIM & 92.70 & 100.00 & 100.00 & 100.00 \\
& DCMorph & 69.20 & 100.00 & 100.00 & 99.80 \\
\bottomrule
\end{tabular}}
\label{tab:mad_results_ablation}
\vspace{-4mm}
\end{table}

\vspace{-1mm}
\subsection{Comparison with SOTA Methods}
\vspace{-1mm}
\begin{table*}[t]
\centering
\caption{Vulnerability comparison of four SOTA FRs to DCMorph and six baseline morphing attacks. Attacks are categorized as image-level (OpenCV, FaceMorpher, WebMorph) or representation-level (MIPGAN-I, MIPGAN-II, MorDIFF, DCMorph). Higher MMPMR values indicate stronger attacks and greater FR vulnerability. DCMorph achieves the highest MMPMR values (bold) across all FR systems at both operational thresholds (MMPMR100 at FMR 1\% and MMPMR1000 at FMR 0.1\%), demonstrating superior effectiveness compared to all representation-level methods and matching or exceeding the best image-level approaches.} \vspace{-2mm}
\setlength{\tabcolsep}{6pt}
\renewcommand{\arraystretch}{1.2}
\resizebox{0.9\textwidth}{!}{%
\begin{tabular}{c c c c c c c c c c}
\toprule
\multicolumn{2}{c}{\multirow{2}{*}{Morphing technique} }
& \multicolumn{2}{c}{ElasticFace \cite{DBLP:conf/cvpr/BoutrosDKK22}}
& \multicolumn{2}{c}{CurricularFace \cite{DBLP:conf/cvpr/HuangWT0SLLH20}}
& \multicolumn{2}{c}{AdaFace \cite{DBLP:conf/cvpr/Kim0L22}}  
& \multicolumn{2}{c}{ArcFace \cite{DBLP:conf/cvpr/DengGXZ19}} \\
\cmidrule(lr){3-4} \cmidrule(lr){5-6} \cmidrule(lr){7-8} \cmidrule(lr){9-10}
& & MMPMR100 & MMPMR1000 & MMPMR100 & MMPMR1000 & MMPMR100 & MMPMR1000 & MMPMR100 & MMPMR1000 \\
\midrule
\multirow{3}{*}{Image level}
& OpenCV     & \textbf{0.997} & 0.980 & \textbf{0.996} & 0.986 & \textbf{1.000} & \textbf{0.993} & \textbf{0.997} & 0.979\\
& FaceMorpher& 0.962 & 0.913 & 0.970 & 0.935 & 0.973 & 0.948 & 0.958 & 0.920\\
& WebMorph   & 0.990 & \textbf{0.986} & 0.988 & \textbf{0.988} & 0.988 & 0.988 & 0.988 & \textbf{0.988} \\
\midrule
\multirow{4}{*}{Representation level}
& MIPGAN-I   & 0.980 & 0.845 & 0.962 & 0.890 & 0.971 & 0.902 & 0.961 & 0.853 \\
& MIPGAN-II  & 0.953 & 0.778 & 0.953 & 0.832 & 0.965 & 0.868 & 0.953 & 0.818 \\
& MorDIFF & 0.990 & 0.948 & 0.995 & 0.968 & 0.991 & 0.962 & 0.985 & 0.917\\
& DCMorph (our) & \textbf{1.000} & \textbf{0.965} & \textbf{0.999} & \textbf{0.981} & \textbf{1.000} & \textbf{0.995} & \textbf{0.998} & \textbf{0.982} \\
\bottomrule
\end{tabular}}
\label{tab:morph_results_sota}
\vspace{-1mm}
\end{table*}

\begin{table}[ht!]
\centering
\caption{Detectability comparison of DCMorph and six baseline morphing attacks across three MAD systems in a realistic cross-dataset evaluation protocol. The table shows EER and APCER at three BPCER operating points (1\%, 10\%, 20\%). Lower values indicate easier detection (better MAD performance). DCMorph demonstrates the most challenging detectability profile across all three MADs: highest EER for MADPromptS, and extremely high EER for SPL-MAD and MixFaceNet-MAD, substantially exceeding all baseline attacks including the recent MorDIFF.} \vspace{-2mm}
\resizebox{0.45\textwidth}{!}{
\begin{tabular}{llcccc}
\toprule
\multirow{2}{*}{Method} & \multirow{2}{*}{Test data}
& \multirow{2}{*}{EER (\%)}
& \multicolumn{3}{c}{APCER (\%) @ BPCER (\%)} \\
\cmidrule(lr){4-6}
& & & 1.00 & 10.00 & 20.00 \\
\midrule
\multirow{7}{*}{MADPromptS \cite{DBLP:journals/corr/abs-2508-08939}}
& FaceMorph  & 18.10 & 61.20 & 25.10 & 16.40  \\
& MIPGAN\_I  & 5.40  & 26.50 & 4.50  & 1.60 \\
& MIPGAN\_II & 3.50  & 13.41 & 1.20  & 0.20 \\
& OpenCV     & 16.06 & 66.97 & 21.14 & 10.98  \\
& WebMorph   & 18.40 & 75.60 & 30.60 & 17.40  \\
& MorDIFF    & 24.50 & 94.70 & 52.20 & 30.10  \\
& DCMorph (our) & 27.60 & 92.50 & 48.90 & 35.70 \\
\midrule
\multirow{7}{*}{SPL-MAD \cite{fang2022unsupervised}}
& FaceMorph  & 0.00 & 65.80 & 65.80 & 0.00\\
& MIPGAN\_I  & 16.30 & 67.30 & 23.00 & 11.20 \\
& MIPGAN\_II & 11.01 & 54.35 & 14.31 & 6.41 \\
& OpenCV     & 5.89 & 20.53 & 3.15 & 1.32 \\
& WebMorph   & 11.60 & 50.00 & 13.00 & 6.00 \\
& MorDIFF    & 7.70 & 25.50 & 6.40 & 2.50 \\
& DCMorph (our) & 77.60 & 100.0 & 100.0 & 100.0 \\
\midrule
\multirow{7}{*}{MixFaceNet-MAD \cite{damer2022privacy}}
& FaceMorph  & 4.60 & 5.60 & 3.70 & 2.90 \\
& MIPGAN\_I  & 16.60 & 75.80 & 22.40 & 14.50 \\
& MIPGAN\_II & 20.52 & 81.68 & 32.13 & 20.62\\
& OpenCV     & 8.33 & 36.48 & 6.50 & 3.86\\
& WebMorph   & 18.20 & 74.20 & 24.00 & 17.60\\
& MorDIFF    & 9.40 & 36.30 & 8.90 & 5.20 \\
& DCMorph (our) & 69.20 & 100.00 & 100.00 & 99.80 \\
\bottomrule
\end{tabular}}
\label{tab:mad_results_sota}
\vspace{-2mm}
\end{table}

\textbf{Vulnerability Analysis:}
We evaluated the vulnerability of four FR systems to DCMorph attacks compared against six different approaches: three image-level methods (OpenCV, FaceMorpher, WebMorph) and three representation-level methods (MIPGAN-I, MIPGAN-II, MorDIFF). Table \ref{tab:morph_results_sota} presents comprehensive vulnerability results across all attack types and FR systems. On all four FR systems at FMR 1\%, DCMorph demonstrates superior attack effectiveness compared to all representation-level baseline attacks. At FMR 1\%, DCMorph achieves attack effectiveness that is competitive with the strongest representation level methods and, for some systems such as AdaFace, attains the highest MMPMR values. While WebMorph yields slightly higher scores in certain cases, DCMorph remains consistently among the top performing approaches across all FR systems. When compared to the diffusion-based MorDIFF, DCMorph shows consistent improvements, achieving an MMPMR at FMR 0.1\% of 0.965 vs. 0.948 on ElasticFace, 0.981 vs. 0.968 on CurricularFace, 0.995 vs. 0.962 on AdaFace, and 0.982 vs. 0.917 on ArcFace.


\textbf{Detectability Analysis:}
Table \ref{tab:mad_results_sota} presents the detectability of DCMorph attacks compared to six different morphing methods across three MAD systems, presented in Section \ref{sec:experimentalsetup}. The evaluation follows a realistic cross-dataset protocol where all MADs are tested on attacks generated from datasets different from their training data, representing practical deployment scenarios. The detectability results reveal varied patterns across different MAD architectures. For MADPromptS, a zero-shot learning approach based on foundation models, DCMorph shows the highest EER (27.60\%) among all attack types, making it the most challenging to detect. At 10\% BPCER, DCMorph achieves an APCER of 48.90\%, substantially higher than image-level methods (21.14–30.60\%) and comparable to MorDIFF (52.20\%). For SPL-MAD, an unsupervised anomaly detection method, DCMorph presents extreme detection challenges with an EER of 77.60\% and APCER of 100\% across all BPCER operating points. This represents the poorest detection performance across all attack types evaluated, far exceeding the detection difficulty of other representation-level attacks (MIPGAN-I: 16.30\% EER, MIPGAN-II: 11.01\% EER, MorDIFF: 7.70\% EER) and even image-level attacks (5.89–11.60\% EER). The extremely high EER suggests that the unsupervised anomaly detector's learned representations fail to distinguish DCMorph attacks from bona fide samples. Similarly, for MixFaceNet-MAD, a supervised detection approach, DCMorph demonstrates substantial detection challenges with an EER of 69.20\% and APCER of 100\% at both 1\% and 10\% BPCER. This is considerably higher than the detection difficulty of other attacks, including MorDIFF (9.40\% EER), image-level methods (4.60–18.20\% EER), and GAN-based methods (16.60–20.52\% EER).

These detectability results, combined with the vulnerability analyses, demonstrate that DCMorph achieves a favorable trade-off from an attacker's perspective: maximizing FR vulnerability while remaining highly challenging to detect by current MAD solutions. The detection difficulty is particularly pronounced for learning-based MADs (SPL-MAD and MixFaceNet-MAD), suggesting that DCMorph's dual-stream approach produces morphs with characteristics that differ from the attack patterns these detectors have been trained to recognize. Only the zero-shot foundation model approach (MADPromptS) shows moderate detection capability, though DCMorph remains more challenging to detect than most analyzed attacks even for this detector.

\vspace{-2mm}
\section{Conclusion}
\vspace{-2mm}
\label{sec:conclusion}
This paper presented DCMorph, a dual-stream morphing framework addressing limitations of existing approaches by operating at both identity conditioning and latent space levels. The framework introduces two key mechanisms: (1) decoupled cross-attention interpolation that injects identity-specific features from both source faces into the denoising process, enabling explicit dual-identity conditioning absent in prior diffusion-based methods, and (2) DDIM inversion with spherical interpolation between inverted latent representations, providing geometrically consistent initialization that preserves structural attributes. Comprehensive analyses across four SOTA FRs show that DCMorph achieves MMPMR values of 0.965–0.995 at FMR 0.1\%, outperforming representation-level baselines, including GAN-based methods (MIPGAN-I, MIPGAN-II) and MorDIFF, while matching or exceeding image-level techniques (OpenCV, FaceMorpher, WebMorph). Ablation studies confirmed that the dual-stream approach consistently outperforms single-stream variants. Detectability analyses across three MAD systems show strong resistance, with EER values of 77.60\% for SPL-MAD and 69.20\% for MixFaceNet-MAD, indicating morphs that evade conventional detection. Cross-dataset evaluation further demonstrates consistent detectability characteristics compared to baseline methods.

\section*{Acknowledgment}
This research work has been funded by the German Federal Ministry of Education and Research and the Hessen State Ministry for Higher Education, Research and the Arts within their joint support of the National Research Center for Applied Cybersecurity ATHENE.

{
    \small
    \bibliographystyle{ieeenat_fullname}
    \bibliography{main}
}


\end{document}